\pdfoutput=1

\documentclass[11pt]{article}

\usepackage{ACL2023}

\usepackage{times}
\usepackage{latexsym}
\usepackage{dsfont}
\usepackage{bm}
\usepackage{amsmath}
\usepackage{amssymb}
\usepackage{booktabs}
\usepackage{multirow}
\usepackage{cellspace}
\usepackage{graphicx}
\usepackage{float}
\usepackage{xspace}
\usepackage{mathtools, cuted}
\usepackage[capitalize,noabbrev]{cleveref}
\mathchardef\mhyphen="2D

\usepackage[T1]{fontenc}

\usepackage[utf8]{inputenc}

\usepackage{microtype}
\usepackage{adjustbox}
\usepackage{tikz}
\usetikzlibrary{positioning,decorations,arrows,shapes,arrows.meta,calc,scopes,shadows,fadings}
\usepackage{tikzscale}

\newcommand{\norm}[1]{\left\lVert #1 \right\rVert}
\newcommand{\abs}[1]{\left\lvert #1 \right\rvert}
\newcommand{\sfA}{{\sf A}}
\newcommand{\sfB}{{\sf B}}
\newcommand{\sfE}{{\sf E}}
\newcommand{\sfC}{{\sf C}}
\newcommand{\sfX}{{\sf X}}

\newcommand{\calX}{{\cal X}}
\newcommand{\calA}{{\cal A}}
\newcommand{\Exp}{\mathbb{E}}
\newcommand{\x}{{\boldsymbol x}}
\newcommand{\ate}[0]{\emph{average treatment effect}\xspace}
\newcommand{\modelname}[0]{\texttt{\textbf{COLA}}\xspace}
\newcommand{\dataname}[0]{\texttt{\textbf{COPES}}\xspace}
\renewcommand{\Pr}{\mathbb{P}}

\title{\modelname: Contextualized Commonsense Causal Reasoning \\from the Causal Inference Perspective}

\author{Zhaowei Wang$^1$, Quyet V. Do$^1$, Hongming Zhang$^2$, Jiayao Zhang$^3$,\\\textbf{Weiqi Wang$^1$, Tianqing Fang$^1$, Yangqiu Song$^1$, Ginny Y. Wong$^4$, \& Simon See$^4$}\\
$^1$Department of Computer Science and Engineering, HKUST\\
$^2$Tencent AI Lab, Bellevue, USA
$^3$University of Pennsylvania\\
$^4$NVIDIA AI Technology Center (NVAITC), NVIDIA, Santa Clara, USA\\
\texttt{\{zwanggy,yqsong\}@cse.ust.hk,} \texttt{\{gwong,ssee\}@nvidia.com}}

\begin{document}
\maketitle
\begin{abstract}
Detecting commonsense causal relations (causation) between events has long been an essential yet challenging task. Given that events are complicated, an event may have different causes under various contexts. Thus, exploiting context plays an essential role in detecting causal relations. Meanwhile, previous works about commonsense causation only consider two events and ignore their context, simplifying the task formulation. This paper proposes a new task to detect commonsense causation between two events in an event sequence (i.e., context), called contextualized commonsense causal reasoning. We also design a zero-shot framework: \modelname (\textbf{C}ontextualized C\textbf{o}mmonsense Causa\textbf{l}ity Re\textbf{a}soner) to solve the task from the causal inference perspective. This framework obtains rich incidental supervision from temporality and balances covariates from multiple timestamps to remove confounding effects. Our extensive experiments show that \modelname\footnote{The code and data are available at 
\url{https://github.com/HKUST-KnowComp/COLA}.} can detect commonsense causality more accurately than baselines.
\end{abstract}

\section{Introduction}
\label{sec:introduction}
Commonsense Causal Reasoning (CCR) aims at identifying plausible causes and effects of events in natural language that are typically reasonable by an average person~\cite{zhang2022rock}. To solve the task, existing efforts devoted by the community mainly rely on language models wholeheartedly with supervised learning approaches~\cite{staliunaite2021improving, sap2019social, tamborrino2020pre, he2020deberta, raffel2020exploring}. Those ingenious engineering works have brought significant progress in recent years. However, recent studies~\cite{kavumba2019choosing, han2021doing} found that pure engineering designs are inadequate to seize commonsense causation, as language models tend to reach higher scores by exploiting superficial artifacts in data.

\begin{figure}[t]
    \centering
    \includegraphics[width=\columnwidth]{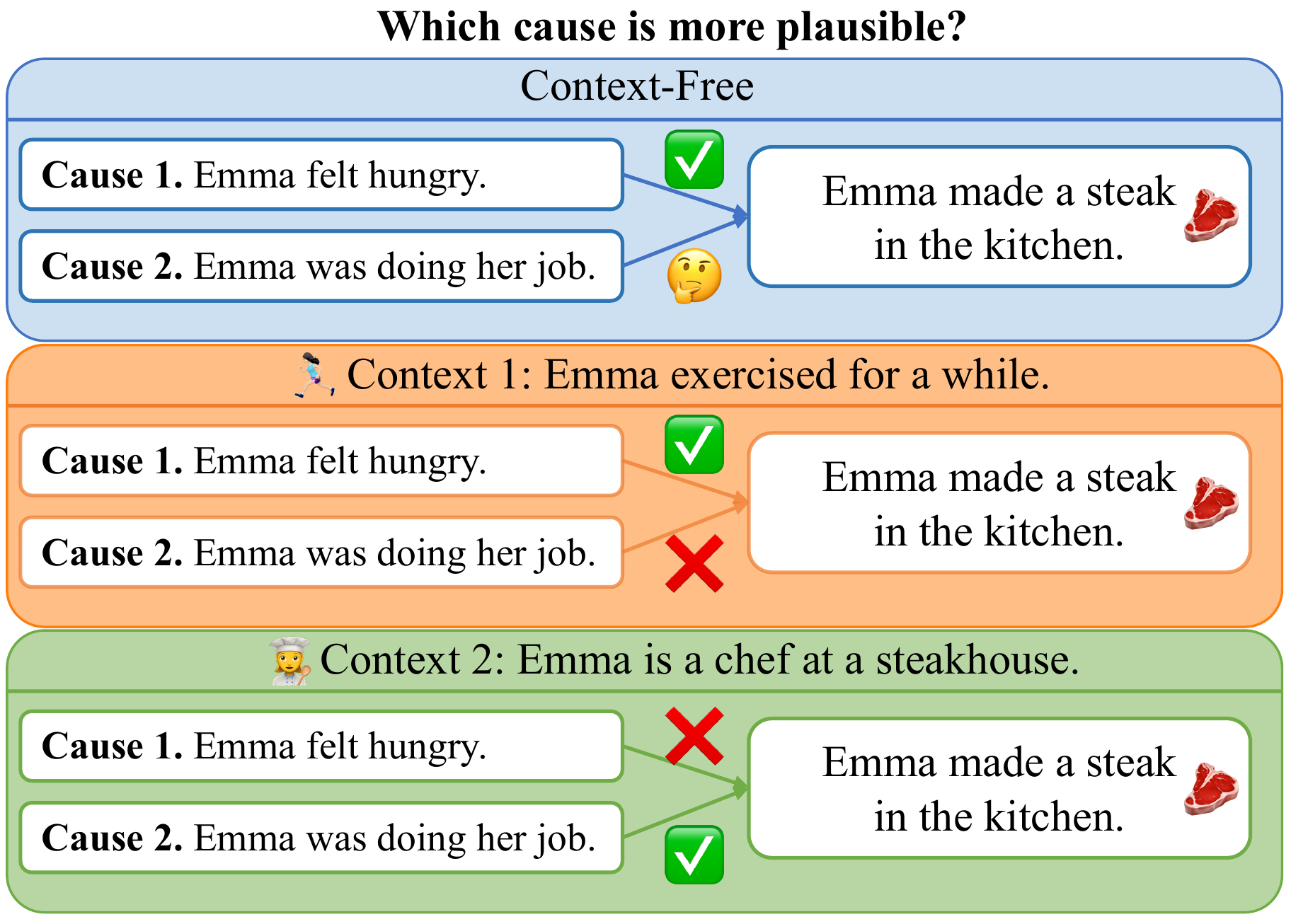}
    \caption{An illustration of leveraging context to conduct commonsense causal reasoning. Both causes could be plausible under different contexts, while only the frequent one (i.e., \textbf{Cause 1}) is plausible without context.}
    \label{fig:intro}
\end{figure}

Recently, \newcite{zhang2022rock} first studied grasping commonsense causation from the causal inference perspective, by drawing analogies between observational studies~\cite{cochran1965planning,rosenbaum2002observational} and natural languages~\cite{zhang2021some}. The proposed framework ROCK achieves good potential for the zero-shot CCR task (e.g., COPA by~\citet{gordon2012semeval}). However, \citet{zhang2022rock} only focuses on the commonsense causation between a pair of events without specifying context. Given that events are complex~\cite{chen2021event}, an event may have different causes under different contexts~\cite{mostafazadeh2020glucose}. Thus, it is necessary to utilize context when detecting commonsense causation, such as other events related to given ones. Missing a clear and specific context simplifies commonsense causal knowledge and hinders models from detecting commonsense causal relations more accurately. 
For example, as shown in \cref{fig:intro}, the frequent cause of ``Emma made a steak in the kitchen.'' is ``Emma felt hungry.'' However, the cause also could be ``Emma was doing her job'' if ``Emma is a chef at a steakhouse.'' Without the context of Emma's job, models cannot distinguish those two causes and may return to the frequent one.

To involve context when detecting commonsense causation, we propose a new task to detect causes between two events in an event sequence, called Contextualized Commonsense Causal Reasoning (Contextualized CCR). In this task, models are asked to detect commonsense causal relations between two given events enclosed in an event sequence. Other events in the event sequence can provide a clear and specific definition of the current context, helping models to capture commonsense causation more accurately. In fact, we find that contextualized CCR is a non-trivial task. Directly applying the framework ROCK~\cite{zhang2022rock} cannot achieve 
competitive performance on the contextualized CCR since it cannot integrate context information.

We propose the framework \modelname, which incorporates contextual information from an event sequence, to solve the Contextualized CCR.
Our framework adopts the potential-outcomes framework~\cite{rubin1974estimating,rosenbaum2002observational,rubin2005causal} to estimate the causal
estimand $\Delta$ defined as a type of
``\ate'' (ATE), which measures the change in the likelihood of $\sfE_j$'s occurrence when intervening $\sfE_i$ (denoted by $\neg \sfE_i$) as
\begin{equation}
\label{eq:ate}
\begin{gathered}
\Delta = \mathbb{P} \left( \sfE_i \prec \sfE_j \right)
	-
	\Pr \left( \neg \sfE_i \prec \sfE_j \right),
\end{gathered}
\end{equation}
where $\Pr(\cdot)$ can be estimated with a pre-trained language model, such as a masked language model~\cite{devlin2018bert}. The magnitude of \ate informs the strength of $\sfE_i$'s effect on $\sfE_j$, and its sign indicates the direction of the effect. For instance, $\Delta \approx 1$ means $\sfE_j$ becomes more prone to occur due to the occurrences of $\sfE_i$. 
In an ideal world (e.g., $\sfE_i$ and $\sfE_j$ on any study unit occur completely randomly), a plugging-in estimator in \cref{eq:ate} suffices for detecting commonsense causation. Nevertheless, spurious correlations introduced by pervasive confounding co-occurrences need to be
eliminated for an unbiased estimation of the causal estimand. This
can be done by \emph{balancing} events that precede $\sfE_i$, or \emph{covariates}. To incorporate context, we design a mechanism to sample  diversified covariates from multiple timestamps and use \emph{temporal propensity}~\cite{zhang2022rock} for balancing.

We annotated commonsense causal relations between two events (\textasciitilde 1.3k examples) within event sequences from ROCStories~\cite{mostafazadeh2016corpus} to benchmark our proposed contextualized CCR task. We conduct extensive experiments with multiple pre-trained language models, showing that \modelname can detect cause-and-effect relations more accurately than competitive baselines by a large margin. Our experiments also show that temporality is essential in our framework but not sufficient 
to detect commonsense causation without covariates being appropriately balanced.

\section{Background and Related Works}
Understanding events and relations between them have long been a challenging NLP task~\cite{chen2021event}. The community has dedicated many works to studying various event-centric tasks, including event relation reasoning~\cite{ning2018multi, zhou2021temporal, wang2020joint}, event extraction~\cite{huang2018zero, lai2020extensively, zhang-etal-2022-efficient-zero, lin2023global}, event-centric KG construction~\cite{zhang2020aser, zhang2022aser}, and many others~\cite{chambers2008unsupervised, chen2020you, jin2022probing, wang-etal-2022-subeventwriter}. Among them, there are a few lines of work that are most related to our work:
\paragraph{Commonsense Causal Reasoning}
Since our work is about Contextualized CCR, we first discuss related works about commonsense causal reasoning.
Existing commonsense causal reasoning approaches are typically categorized under the general topic of commonsense reasoning~\cite{rashkin2018event2mind, sap2020commonsense}. Most previous works depend on language models. Remarkable progress in CCR mainly comes from dataset augmentation, training procedure design, and external knowledge~\cite{staliunaite2021improving, sap2019social, shwartz2020unsupervised, tamborrino2020pre, iter2020pretraining}. Studies~\cite{kavumba2019choosing, han2021doing} show that language models exploit superficial artifacts to achieve suspicious high performance. 

Causal event detection~\cite{mirza2014analysis, mirza2014annotating} forms another line of work pertinent to CCR. The task aims to detect causal relations in documents, where various methods are proposed~\cite{chang2005causal, do2011minimally, ning2019joint}. However, those works consider verbal (e.g., “attack”) or nominal predicates (e.g., “explosion”) as events, oversimplifying the relation detection task. In this paper, we study events expressed in free text, facing a more challenging setup but being closer to real applications.

\paragraph{Narrative-related Tasks} Since Contextualized CCR is primarily about chains of events, our work is inspired by earlier research that deals with narratives and scripts~\cite{chambers2008unsupervised, granroth2016happens, mostafazadeh2016corpus, bhagavatula2019abductive, zhang2020analogous}. In contrast, our work aims to identify causal relations in a chain of events.

\paragraph{Methodologies of Causal Inference}
\citet{zhang2021some} provided the first study to solve the CCR task from the causal inference perspective. The human population studies have scrutinized extensively the causal inference, which identifies causal relations from ubiquitous associations, including biomedical research, agriculture, epidemiology, and economics~\cite{fisher1958cancer, imbens2015causal, giannarakis2022towards, rosenbaum2002observational, cochran1965planning}, where researchers usually use the potential-outcomes framework~\cite{splawa1990application, rubin1974estimating, holland1986statistics}, graphical and structural equation models~\cite{robins1986new, pearl1995causal, heckman2005rejoinder}, and Granger causality~\cite{granger1969investigating}.

Recent studies have drawn causal inferences on textual data with the help of powerful pre-trained language models~\cite{kang2017detecting, keith2020text, feder2022causal}. Concurrently, causal inference can improve the robustness and fairness of NLP models~\cite{feder2022causal} or boost performance on downstream tasks~\cite{ghosal2021cider, zheng2022distilling, alabdulkarim2021automatic, wang2022open}.

\begin{figure*}[t]
    \centering
    \includegraphics[width=2\columnwidth]{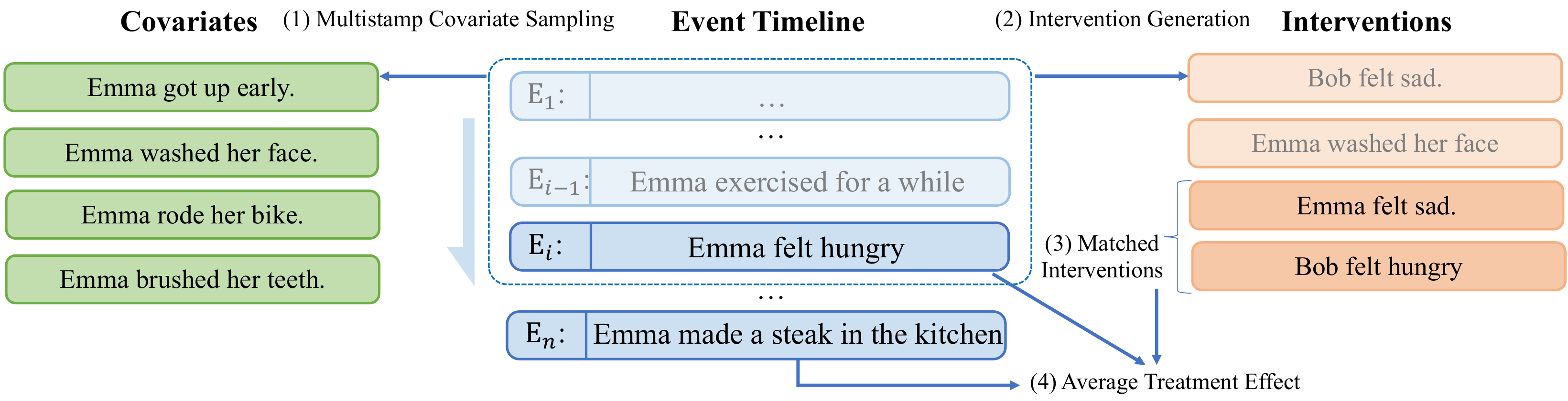}
    \caption{The overview of our proposed framework \modelname. Given a pair of events ($\sfE_i, \sfE_n$), the framework contains four steps to detect commonsense causal relations: (1) multistamp covariate sampling, (2) intervention generation (3) selecting a set of matched interventions, (4) computing \ate.}
    \label{fig:method_figure}
\end{figure*}

\section{Problem Formulation}
\label{sec:problem_formulation}
\paragraph{Notation} We use $\sf{sans-serif}$ fonts to represent an event, such as $\sfE_i$ in \cref{fig:method_figure}, where the subscript $i$ means the $i$-th event in a sequence. Simultaneously, uppercase serif letters denote \emph{indicators} of whether the corresponding event occurs: $E_i = \mathds{1}\left\{\sfE_i \text{ occurs}\right\}$, and a lowercase serif letter means the realizations of this indicator: $e_{i,j} =
\mathds{1}\left\{\sfE_i \text{ occurs to the }j\text{-th study unit}\right\}$. We introduce the point process to more clearly describe the order of \textbf{a pair of} events: $E(t)$ with $t \in \{0, 1\}$ (e.g.,
past versus present), so that $E_i(0)$ and $E_j(1)$ means that $\sfE_i$ happens before $\sfE_j$. We also use $\sfE_i \prec \sfE_j$ to indicate that $\sfE_i$ occurs before $\sfE_j$ for simplicity. We write $\Pr(\sfE_i\prec \sfE_j) = \Pr(E_i(0), E_j(1))$.

\paragraph{Task Description}
We articulate the Contextualized CCR as a tweaked form of the binary classification problem. Specifically, we provide models with an event sequence of $n$ events: $\sfE_1, \sfE_2, \ldots, \sfE_n$. Models need to find the top-$k$ events in this sequence that more plausibly have commonsense causal relations with the last event $\sfE_n$, where $k$ indicates the number of positive labels in the ground truth. Then, models need to predict the commonsense causal relation between each event pair ($\sfE_i$, $\sfE_n$) as a positive/negative one. The strength of causal relations between $\sfE_i$ and $\sfE_n$ can be expressed with \ate as:
\begin{equation}
\label{eq:task_definition}
\begin{gathered}
\Delta_i = \mathbb{P} \left( \sfE_i \prec \sfE_n \right)
	- 
	\Pr \left( \neg \sfE_i \prec \sfE_n \right).
\end{gathered}
\end{equation}
\section{Theoretical Mechanism of \modelname}
\label{sec:theoretical_mechanism}
As discussed in \cref{sec:introduction}, we articulate the Contextualized CCR problem as the estimation of the causal estimand $\Delta$, which
we model as the change of temporal likelihood \emph{with contexts controlled}.
We adopt the potential-outcomes framework to design \modelname to eliminate potential
confounding effects due to co-occurrences of events when estimating the causal estimand from data.
In this section, we first clarify all essential concepts in the theoretical mechanism of \modelname one by one, including \emph{study unit}, \emph{intervention}, \emph{covariate}, and \emph{propensity}, by
drawing analogies between the underlying causal mechanisms in natural
languages with that in human population research. We then describe the implementation of each component in \cref{sec:the_cola_framework}.

\subsection{The Analogy and Study Unit}\label{sec:building_parallel}
Motivated by~\citet{zhang2022rock}, we draw the analogy between human subjects and semantic meanings through the following process: assuming that every human subject kept a textbook recording each event (s)he has experienced, we can then treat each textbook (in natural language) as a study unit and infer the temporal relationships between events from it. In this analogy, we clearly understand the study unit in semantic meanings. 

Then, we can formulate contextualized CCR
with concepts from the potential-outcome framework. Given two events $\sfE_i$ and $\sfE_n$ from an event sequence $\sfE_1, \sfE_2, \ldots, \sfE_n$, where we assume that $\sfE_{ij}$ represents the event that the $j$-th study unit experienced at the timestamp $i$ when $\sfE_i$ is supposed to occur. Then for each unit $j$, we can define the treatment assignment as
$E_{ij} = \mathds{1}\left\{\sfE_{ij} = \sfE_i \right\}$, realizations of covariates as $\x_j = (x_{jl})_{l=1}^N$ for $x_{jl} = \mathds{1}\left\{\sfX_l\prec \sfE_{ij}\right\}$, and two potential-outcomes as
\begin{equation}\begin{aligned}\begin{cases}
        r_{0, ij} = \mathds{1}\left\{\sfE_{ij,E_{i}=0} \prec \sfE_n\right\}, \\
        r_{1, ij} = \mathds{1}\left\{\sfE_{ij,E_{i}=1} \prec \sfE_n\right\}.
\end{cases}\end{aligned}\end{equation}
When the $j$-th unit receives the treatment assignment $E_{ij}$, the hypothetical scenario is denoted by $\sfE_{ij,E_{i}=1-E_{ij}}$, which describes what if the assignment were flipped. Clearly, we can only observe either $r_{0,ij}$ or $r_{1,ij}$, but not both of them. We can rewrite the causal estimand $\Delta_i$ in \cref{eq:task_definition} exactly as an \ate by averaging over the unit index:
\begin{equation}
    \begin{aligned}
        \Delta_i &= \Exp_j[r_{1,ij} - r_{0,ij}] \\
        &\equiv \Pr(\sfE_i \prec \sfE_n)
        - \Pr(\neg \sfE_i \prec \sfE_n).
    \end{aligned}
\end{equation}
The above formulation naturally embodies the temporal nature of 
 covariates~\cite{rubin2005causal}, which, by definition, are pretreatments that precede treatments.

\subsection{Intervention Beyond Negation}
\label{sec:intervention_definition}
In human population studies, the generally accepted \emph{stable unit treatment value assumption}~\cite{rubin1980bias} ensures only one type of non-treatment (usually negation) for each study unit. As events are complicated, we would interpret intervention (manipulation) of semantic meanings in a broader sense. Take $\sfE_i$ ``Emma felt hungry'' from \cref{fig:method_figure} as an example. While ``Emma didn't feel hungry'' is the most direct intervention, it is nonetheless too restrictive: Emma may have felt happy; maybe Alice is the one who felt hungry, instead of Emma.
Consequently, interventions in our framework are interpreted much broader as any event that could result in a plausible counterfactual of the outcome. We use $\calA$ to represent all possible interventions of an event $\sfE_i$.

\subsection{Balancing Covariates and Comparable Study Units}
We have discussed that the plugging-in estimator in \cref{eq:ate} suffers from biases due to potential confounders. One mitigation strategy is properly balancing the covariates~\cite{rubin2005causal}, namely events that occur before $\sfE_i$, which ensures that covariates of untreated study units are comparable to those of treated ones. Consider the vaccine trial as an example; one needs to ensure that the health conditions (covariates) in the control group (not administered vaccines) are comparable to the treated group (administered vaccines).
As such, we rewrite the causal estimand
$\Delta$ in \cref{eq:task_definition} as expectations conditional on the covariates $\x$ among comparable study units:
\begin{equation}
    \label{eq:conditional_ate}
    \begin{aligned}
	\Exp_{\x} \left[ \Pr(\sfE_i \prec \sfE_n \vert \x)
		- \Pr(\neg \sfE_i \prec \sfE_n \vert \x) \right],
    \end{aligned}
\end{equation}
provided that the treatment assignment is strongly ignorable with respect to potential outcomes (i.e., $r_{1,ij}$ and $r_{0,ij}$) according to the strong ignorability assumption. 

The strong ignorability assumption is essential in causal inference. Simply, it means that given a set of covariates,  the treatment
assignment among study units can be viewed as ``random'' (or ``ignorable'')
with respect to potential outcomes (see, e.g., \citet{rosenbaum2002observational,rubin2005causal} for textbook treatments
and \citet{zhang2022rock} for more discussions on this assumption
in CCR).

\subsection{Matching Temporal Propensities}
Directly estimating \cref{eq:conditional_ate} may face the issue of data sparsity since we may sample multiple covariates, and combinations of their values grow exponentially.
There are various methods to mitigate this issue in balancing covariates, including assignment modeling, outcome modeling, and doubly-robust estimations \cite{rubin2005causal}, among which we base our method on propensity score
matching. It is a simple and effective method that is widely used
in observational studies~\cite{rosenbaum1983central}, which matches the propensity scores of study units to balance covariates. The propensity score is defined as
$p(\x) = \Pr(E_i(1) = 1\vert \x(0))$,
which represents the probability of $\sfE_i$ taking place conditioning on covariates $\x$. 
Since it is unclear how to pack an unordered set of covariates (events) into a sequential input, \citet{zhang2022rock} proposed to use a relaxed notion
of temporal propensity vector, defined as the vector of probabilities of $\sfE_i$ happening conditional on each covariate $x\in\x$:
\begin{equation}
    \begin{aligned}
       q(\x)=q(\x;\sfE_i) = \left( \Pr(E_i(1) = 1\vert x(0))\right)_{x\in\x}.
    \end{aligned}
\end{equation}
Hence, we can rewrite the conditional expectation in \cref{eq:conditional_ate} in the form of matching temporal propensity vectors for some fixed threshold $\epsilon$, given below:
\begin{equation}
    \label{eq:estimated_ate}
    \begin{aligned}
        \begin{cases}
        \hat{\Delta}_{i}  =
	    f(\sfE_i,\sfE_n) - \frac{1}{\abs{\calA'}}
		\sum_{\sfA \in \calA'} f(\sfA,\sfE_n), \\
	\calA' \coloneqq
	\resizebox{0.7\columnwidth}{!}{%
	$\left\{\sfA \in \calA:
    \frac{1}{\abs{\calX}}\norm{q(\x; \sfA) - q(\x;\sfE_i)}_2 \le \epsilon \right\}$},
        \end{cases}
    \end{aligned}
\end{equation}
where $f(\sfE_i, \sfE_n)$ is an estimate for $\Pr(\sfE_i \prec \sfE_n)$ produced by a language model.

\section{The \modelname Framework}
\label{sec:the_cola_framework}
After establishing the theoretical mechanism for our framework \modelname, we describe the implementation of each component of \modelname in this section. Generally, since events are in free-text form in our task, pre-trained language models play a central role in our framework. Given that LMs are pre-trained on an enormous amount of textual data~\cite{gao2020pile, raffel2020exploring}, it is sensible to suppose that those LMs would emulate the responses of an average reasonable person.

Specifically, our framework \modelname takes two events $\sfE_i$ and $\sfE_n$ from a sequence $\sfE_1, \sfE_2, \ldots, \sfE_n$ as input. As shown in Figure~\ref{fig:method_figure}, our framework \modelname contains four steps: (1) a multistamp covariate sampler samples a set $\calX$ of covariates. (2) an intervention generator generates a set $\calA$ of interventions. (3) A score estimator builds temporal propensity vectors and selects a matched subset $\calA'$ out of $\calA$, by estimating the temporality with a temporal predictor. (4) Eventually, the same score estimator computes $\hat{\Delta}_i$ according to \cref{eq:estimated_ate}.

\paragraph{Multistamp Covariate Sampler}
Our multistamp covariate sampler is based on GPT-J (6b)~\cite{gpt-j}. For an input event $\sfE$, we add ``Before that,'' at the end to build ``$\sfE$ Before that,'' as the prompt template. For the $i$-th event in a sequence, we first sample a covariate set $\calX_i$, which contains events before $\sfE_i$. To diversify covariates, we also sample events before $\sfE_1, \sfE_2, \ldots, \sfE_{i-1}$ separately, forming $\calX_1, \calX_2, \ldots, \calX_{i-1}$. Those events are also before $\sfE_i$ and can serve as covariates due to the transitivity of the temporal relation\footnote{In a sequence of temporally ordered events if $\sfA \prec \sfB$ and $\sfB \prec \sfC$, then $\sfA \prec \sfC$.}. We evenly merge covariates before each timestamp to construct the final covariate set:
\begin{equation}
    \begin{aligned}
       \calX = \cup_{l=1}^{i} \calX_{l}.
    \end{aligned}
\end{equation}

\paragraph{Union vs. Intersection}\label{par:method_union_and_inter} While the aforementioned method takes the union of events sampled according to the left-side context of $\sfE_i$, another intuitive approach is to take the intersection of events sampled according to multiple timestamps in the right-side context. In this way, we can collect covariates that happen before all of $\sfE_i, \sfE_{i+1}, \ldots, \sfE_n$, that is, $\calX = \cap_{l=i}^{n} \calX_{l}$. We discuss the experimental results of these two methods in \cref{sec:union_and_inter} and found taking \textbf{\emph{union}} works better, which will be the default in the remaining sections.

\paragraph{Intervention Generator}
This component generates a set $\calA$ of interventions as discussed in \cref{sec:intervention_definition}. There are a variety of related works about generating interventions (counterfactuals) of an event~\cite{gardner2020evaluating, qin2019counterfactual, ribeiro2020beyond} and we choose \texttt{PolyJuice}~\cite{wu2021polyjuice} in our framework owing to its non-task-specific training objective. \texttt{PolyJuice} generates interventions by masking some phrases individually and filling in masks with a fine-tuned GPT2. Then, we apply the semantic role labeling (SRL) tool provided by AllenNLP~\cite{gardner2018allennlp} to extract the verb ${\tt V}$ and two arguments ${\tt ARG0}$ and ${\tt ARG1}$ as phrases to be manipulated (see \cref{app:controle_code} for more details).

\paragraph{Temporal Predictor}\label{par:method_temporal_predictor}
We prompt a masked language model to estimate the temporal relation scores between two given events $\sfE_i$ and $\sfE_n$. The prompt template ``$\sfE_i \text{ \tt   <MASK> } \sfE_n$'' predicts scores $f_b(\sfE_i, \sfE_n)$ and $f_a(\sfE_i, \sfE_n)$ for the output tokens \texttt{before} and \texttt{after}. Similarly, we can obtain a reversed estimation by inputting ``$\sfE_n \text{ \tt   <MASK> } \sfE_i$.'' Final temporal score $f$ averages scores from both directions: $f(\sfE_i, \sfE_n)= \frac{1}{2}(f_b(\sfE_i, \sfE_n) + f_a(\sfE_n, \sfE_i))$

Our temporal predictor needs to be fine-tuned on a temporal relation corpus. Directly applying a pre-trained LM encounter the problem of low coverage, where the tokens \texttt{before} and \texttt{after} cannot be found in the top-$k$ prompted tokens (even $k=30$). Thus, we fine-tuned masked language models to predict the masked connectives in a prompt learning setting. Intuitively, temporal relations exist between each pair of adjacent events in a chronologically ordered event sequence. Assuming an event sequence contains two adjacent events $\sfE_i, \sfE_{i+1}$, we then can create an example $\sfE_i \text{ \tt   before } \sfE_{i+1}$ and an symmetric example $\sfE_{i+1} \text{ \tt   after } \sfE_i$. We also construct negative samples by replacing $\sfE_i$ or $\sfE_{i+1}$ with a randomly sampled event from other sequences. Those negative examples can teach models when no temporal relation exists. While we mask \texttt{before} or \texttt{after} in a positive example for models to predict, a special token \texttt{[none]} should be prompted for negative examples. Then, the cross-entropy loss is used to optimize the temporal predictor. We call this fine-tuning process \textbf{\emph{temporal fine-tuning}}. More details about the temporal relation dataset and fine-tuning process are shown in \cref{app:temporal_predictor}.

\paragraph{Score Estimator}
\label{sec:score_estimator}
With the temporal predictor, we can estimate $\Pr(\sfX \prec \sfA)$ for all covariates $\sfX \in \calX$ and interventions $\sfA\in\calA$: $\Pr(X(0), A(1)) = \Pr(\sfX \prec \sfA) = f(\sfX, \sfA)$. We also need to estimate $\Pr(X(0))$ to compute conditional probabilities $\Pr(A(1) = 1 \vert X(0))$ in temporal propensity vectors $q(\x; \sfA)$. As all covariates $\sfX$ are events preceding $\sfE_i$ sampled by GPT-J, there is an implicit conditioning on $\sfE_i$. Thus, we can approximately get $\Pr(X(0)) \approx f(\sfX, \sfE_i)$ (see \cref{app:probability_approximation} for more details). Then, temporal propensity vectors are computed as
 \begin{equation}
    \label{eq:est_propensity}
     \begin{aligned}
                 q(\x;\sfA) = \left( \frac{\Pr(X(0), A(1))}{\Pr(X(0))}\right)_{\sfX\in\calX}.
     \end{aligned}
 \end{equation}
Finally, the score estimator computes $\hat{\Delta}_i$ in \cref{eq:estimated_ate}. We also test normalization methods in \cref{sec:normalization} and observe that some of those normalization methods can benefit our framework.

\begin{table*}[t]
    \small
	\centering
	\begin{tabular}{l||cccc|cccc}
	\toprule
        \multirow{2}{*}{\textbf{Models}}&\multicolumn{4}{c|}{\textbf{Validation}} &\multicolumn{4}{c}{\textbf{Testing}}\\
	&\textbf{Acc}&\textbf{F1} &\textbf{Ma-F1} &\textbf{$\Delta_{Acc}$}&\textbf{Acc}&\textbf{F1} &\textbf{Ma-F1} & \textbf{$\Delta_{Acc}$} \\
            \midrule
            Random &59.47&42.35&55.55&-&58.94&41.10&54.79&-\\
		\midrule
            CLM Perplexity (GPT2) &61.76&45.61&58.06&-&61.47&44.73&57.58&-\\
            CLM Perplexity (GPT2-medium) &60.29&43.51&56.45&-&61.76&45.15&57.90&- \\
            CLM Perplexity (GPT2-large) &62.94&47.28&59.35&-&62.65&46.41&58.87&- \\
            CLM Perplexity (GPT2-XL) &62.65&46.86&59.03&-&62.35&45.99&58.55&-\\
            CLM Perplexity (GPT-J 6b) &63.82&48.54&60.32&-&62.06&45.57&58.22&-\\
            \midrule
            ClozePromptScore (BERT-base)&64.41&49.37&60.97&-&63.53&47.68&59.84&-\\
            ClozePromptScore (BERT-large)&66.47&52.30&63.23&-&62.06&45.57&58.22&-\\
            ClozePromptScore (RoBERTa-base)&59.71&42.68&55.81&-&59.71&42.19&55.63&-\\
            ClozePromptScore (RoBERTa-large)&60.59&43.93&56.77&-&59.12&41.35&54.99&-\\
            ClozePromptScore (DeBERTa-base)&56.76&38.49&52.58&-&58.53&40.51&54.34&-\\
            ClozePromptScore (DeBERTa-large)&56.47&38.08&52.26&-&57.06&38.40&52.72&-\\
            \midrule
            ROCK (BERT-base)&66.18&51.88&62.90&-&65.29&50.21&61.79&-\\
            ROCK (BERT-large)&65.59&51.05&62.26&-&66.47&51.90&63.08&-\\
            ROCK (RoBERTa-base)&61.76&45.61&58.06&-&61.18&44.30&57.25&-\\
            ROCK (RoBERTa-large)&62.94&47.28&59.35&-&65.59&50.63&62.11&-\\
            ROCK (DeBERTa-base)&62.65&46.86&59.03&-&60.59&43.46&56.61&-\\
            ROCK (DeBERTa-large)&64.41&49.37&60.97&-&64.12&48.52&60.49&-\\
            \midrule
            \modelname(BERT-base)&67.65&53.97&64.52&\multicolumn{1}{l|}{$\uparrow$ 1.47}&68.82&55.27&65.67&\multicolumn{1}{l}{$\uparrow$ 3.53}\\
            \modelname(BERT-large)&70.29&57.74&67.42&\multicolumn{1}{l|}{$\uparrow$ 4.70}&\textbf{70.29}&\textbf{57.38}&\textbf{67.29} &\multicolumn{1}{l}{$\uparrow$ 3.82}\\
            \modelname(RoBERTa-base)&69.71&56.90&66.77&\multicolumn{1}{l|}{$\uparrow$ \textbf{7.95}}&66.76&52.32&63.41&\multicolumn{1}{l}{$\uparrow$ 5.58}\\
            \modelname(RoBERTa-large)&70.59&58.16&67.74&\multicolumn{1}{l|}{$\uparrow$ 7.65}&70.00&56.96&66.97&\multicolumn{1}{l}{$\uparrow$ 4.41}\\
            \modelname(DeBERTa-base)&69.71&56.90&66.77&\multicolumn{1}{l|}{$\uparrow$ 7.06}&\textbf{70.29}&\textbf{57.38}&\textbf{67.29}&\multicolumn{1}{l}{$\uparrow$ \textbf{9.70}}\\
            \modelname(DeBERTa-large)&\textbf{71.18}&\textbf{59.00}&\textbf{68.39}&\multicolumn{1}{l|}{$\uparrow$ 6.77}&69.41&56.12&66.32&\multicolumn{1}{l}{$\uparrow$ 5.29}\\
		\bottomrule
	\end{tabular}
	\caption{Performance of all frameworks on the validation and testing set of the \dataname dataset. \modelname is our model. We abbreviate Accuracy, F1-score, Macro F1-score to Acc, F1, Ma-F1, respectively. We test \modelname with the temporal predictor based on different models: BERT-base/large, RoBERTa-base/large, and DeBERTa-base/large. Compared with ROCK, improvements of our frameworks are shown under \textbf{$\Delta_{Acc}$} for each language model, respectively.}
    \label{tab:main_eval}
\end{table*}

\section{Experiment}
We conduct extensive experiments and compare
\modelname with a wide selection of baselines.
\subsection{Dataset}
Since our work is the first attempt to study Contextualized CCR, we carried out human annotation on Amazon Mechanical Turk.
We randomly sampled event sequences from ROCStories~\cite{mostafazadeh2016corpus}, where each sequence contains five chronologically ordered events. Workers are asked to annotate whether an event causes the last event in a sequence. There are two qualification tests to choose workers to maintain rigorous quality control. See more details in \cref{app:dataset_annotation}.

Eventually, we collected a dataset containing 1,360 event pairs, called \emph{Choice of Plausible Event in Sequence} (\dataname). We equally divide them into a validation set and a testing set.

\subsection{Evaluation Metric}
We calculate accuracy, F1-score, and Macro F1-score between predicted labels and ground truth labels, to automatically evaluate all models on our dataset. Notice that our task definition provides the number of positive events in a sequence, so that recall, precision, and F1-score are the same.

\subsection{Baseline Methods}
We compare our framework to three baselines:

\paragraph{CLM Perplexity} An intuitive solution to
the contextualized CCR task would be computing perplexity scores for each pair of events with a causal language model (CLM). An event pair ($E_i$, $E_j$) within a sequence $E_1, E_2, \ldots, E_n$ ($n$ is sequence length) is converted into full-text input with the prompt template: ``If $E_1, E_2, \ldots, E_n$, $E_j$ because $E_i$''. The causal language models we tested are GPT2, GPT2-medium/large/XL~\cite{radford2019language}, and GPT-J~\cite{gpt-j}.

\paragraph{Cloze Prompt Score} This baseline proposed by~\citet{tamborrino2020pre} concatenates two events ($E_i$, $E_j$) into full-text input. Then, it masks and tries to recover each token with a masked language model. It averages log-likelihood over every token as the final score of two events. The prompt used is the same as \textbf{CLM Perplexity}. Multiple masked language models are tested: BERT-base/large~\cite{devlin2018bert}, RoBERTa-base/large~\cite{liu2019roberta}, DeBERTa-base/large~\cite{he2020deberta}.

\paragraph{ROCK} This baseline is a causal inference framework~\cite{zhang2022rock} that draws analogies between human subjects and natural language. We test different language models for the temporal predictor: BERT-base/large, RoBERTa-base/large, and DeBERTa-base/large.

\begin{figure*}[t]
    \centering
    \includegraphics[width=2\columnwidth]{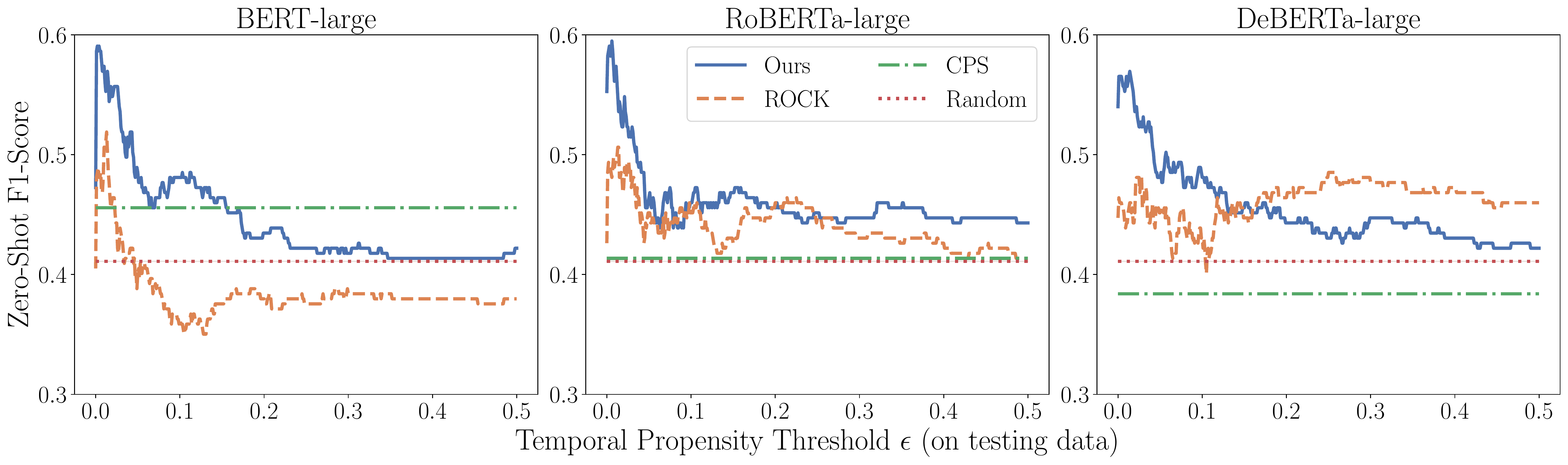}
    \caption{F1-score of \modelname (``Ours'') with $\epsilon$ ranging from 0 to 0.5. We also compare our framework with baselines ClozePromptScore (``CPS'') and ROCK. Balancing covariates in our framework can significantly outperform unbalanced variants ($\epsilon \approx 0.5$).}
    \label{fig:eps}
\end{figure*}

\section{Main Evaluation}
\label{sec:main_evaluation}
We provide results in \cref{tab:main_eval} for baselines and \modelname with the temporal predictor based on different language models. In general, \modelname can detect commonsense causal relations more accurately, outperforming all baseline models by a large margin. Our framework \modelname based on DeBERTa-large and DeBERTa-base (also BERT-large) achieves the best performance on the validation and testing set, respectively. Also, changing language models of the temporal predictor in \modelname only involves a small fluctuation in performance, showing that our framework is robust to underlying language models.

Another observation is that CLM Perplexity and ClozePromptScore can achieve performance higher than the random result. This manifests that pre-trained language models with commonly used pre-training objectives can capture commonsense causal relations to some extent. 

\begin{table}[t]
    \small
	\centering
        \setlength{\tabcolsep}{4.5pt}
	\begin{tabular}{l|cccl}
	\toprule
	\textbf{Models}&\textbf{Acc}&\textbf{F1}&\textbf{Ma-F1}&\multicolumn{1}{c}{\textbf{$\Delta_{Acc}$}}\\
\midrule
Ours(BERT-large)&\textbf{70.29}&\textbf{57.38}&\textbf{67.29}&\multicolumn{1}{c}{-}\\
\midrule
$\diamond$ w/o Multi Step&70.00&56.96&66.97&$\downarrow$ 0.29\\
$\diamond$ w/o Inter&60.29&43.04&56.28&$\downarrow$ 10.00\\
$\diamond$ w/o Cov&57.65&39.24&53.37&$\downarrow$ 12.64\\
$\diamond$ w/o Temp&64.41&48.95&60.82&$\downarrow$ 5.88\\
\midrule
\midrule
Ours(RoBERTa-large)&\textbf{70.00}&\textbf{56.96}&\textbf{66.97}&\multicolumn{1}{c}{-}\\
\midrule
$\diamond$ w/o Multi Step&67.94&54.01&64.70&$\downarrow$ 2.06\\
$\diamond$ w/o Inter&64.71&49.37&61.14&$\downarrow$ 5.29\\
$\diamond$ w/o Cov&58.53&40.51&54.34&$\downarrow$ 11.47\\
$\diamond$ w/o Temp&65.00&49.79&61.46&$\downarrow$ 5.00\\
\midrule
\midrule
Ours(DeBERTa-large)&\textbf{69.41}&\textbf{56.12}&\textbf{66.32}&\multicolumn{1}{c}{-}\\
\midrule
$\diamond$ w/o Multi Step&68.24&54.43&65.03&$\downarrow$ 1.17\\
$\diamond$ w/o Inter&67.65&53.59&64.38&$\downarrow$ 1.76\\
$\diamond$ w/o Cov&55.88&36.71&51.42&$\downarrow$ 13.53\\
$\diamond$ w/o Temp&64.41&48.95&60.82&$\downarrow$ 5.00\\
\bottomrule
\end{tabular}
\caption{Ablation study on \modelname (``Ours''). The column \textbf{$\Delta_{Acc}$} indicates accuracy drops. The first three experiments shows the effectiveness of temporal propensity matching and the last one is about temporal predictor.}
 \label{tab:part_ablation}
\end{table}

\subsection{Ablation Study}
In this section, we conduct four ablation experiments to demonstrate that our causal inference-motivated framework can mitigate spurious correlations between events and boost performance.

\paragraph{Temporal Propensity Matching}
The first three ablation experiments prove the effectiveness of temporal propensity matching. Here, we separately remove three modules in our framework: (i) We drop the \emph{multistamp covariate sampler} and sample covariates only based on the last timestamp ($\diamond$ w/o Multi Step). This experiment verifies the benefit of utilizing context to detect commonsense causality. (ii) We remove all interventions ($\diamond$ w/o Inter) and use temporal precedence as causation: $\hat{\Delta}_i=\Pr(\sfE_i \prec \sfE_n)$, equivalent to $\epsilon=0$ in \cref{eq:estimated_ate}. (iii) Covariates are removed ($\diamond$ w/o Cov) so that interventions are not adjusted. This unadjusted score keeps all sampled interventions, equivalent to $\epsilon=1$ in \cref{eq:estimated_ate}.

From the results in \cref{tab:part_ablation}, we observe that balanced estimand $\hat{\Delta}_i$ achieve better performances, showing that multiple timestamp sampling, treatment effect, and balancing covariates play essential roles in detecting commonsense causation accurately. Removing any of the three modules will result in sheer drops in all metrics. These experiments imply that temporal relation is vital in Contextualized CCR, but it is still insufficient due to spurious correlations. Thus, we need to measure \ate with balancing covariates.

\paragraph{Temporal Predictor}
We also ablate the temporal predictor ($\diamond$ w/o Temp) to verify the effectiveness of \textbf{\emph{temporal fine-tuning}} (in \cref{par:method_temporal_predictor}). Here, we use pre-trained language models and increase $k$ to 30 to mitigate the problem of low coverage.

As shown in \cref{tab:part_ablation}, a directly pre-trained language model without fine-tuning cannot perform well. We conclude that pre-trained language models do not have sufficient ``temporal awareness'' and \textbf{\emph{temporal fine-tuning}} is necessary for our framework.

\subsection{Rules-of-thumb for Choosing $\epsilon$:}
\label{sec:choosing_epsilon}
The hyperparameter $\epsilon$ controls the number of interventions when balancing covariates in \cref{eq:estimated_ate}. In \cref{fig:eps}, we can observe that a recommended range for $\epsilon$ is $\epsilon \in [0, 0.1]$. We also list the optimal $\epsilon$ in \cref{sec:hyperparameter}. From \cref{tab:best_eps}, $\epsilon$ should be fairly small within [0.001, 0.015]. Though the best solution relies on how to implement the components in \modelname and data distribution, our analysis can provide a good starting point. We also study the effect of changing another important hyperparameter: covariate set size $N = \abs{\calX}$, in \cref{app:number_of_covariates}.

\begin{table}[t]
    \small
	\centering
        \setlength{\tabcolsep}{4.5pt}
	\begin{tabular}{l|cccc}
	\toprule
	\textbf{Models}&\textbf{Acc}&\textbf{F1}&\textbf{Ma-F1}&\textbf{$\Delta_{Acc}$}\\
\midrule
\textbf{Uni} (BERT-large)&70.29&57.38&67.29&-\\
\textbf{Int} (BERT-large)&68.82&54.89&65.54&$\downarrow$ 1.47\\
\midrule
\textbf{Uni} (RoBERTa-large)&70.00&56.96&66.97&-\\
\textbf{Int} (RoBERTa-large)&68.82&54.89&65.54&$\downarrow$ 1.18\\
\midrule
\textbf{Uni} (DeBERTa-large)&69.41&56.12&66.32&-\\
\textbf{Int} (DeBERTa-large)&68.82&54.89&65.54&$\downarrow$ 0.59\\
\bottomrule
\end{tabular}
\caption{Comparison of the different covariate sampling methods. \textbf{Uni} and \textbf{Int} stand for ``Union'' and ``Intersection,'' respectively.}
 \label{tab:part_union_inter}
\end{table}

\subsection{Union and Intersection}
\label{sec:union_and_inter}
Our framework \modelname samples covariates from multiple timestamps and take union on them to get the final covariate set. We also introduce another method to sample covariates preceding all timestamps after and including $E_i$ (``intersection'') in \cref{par:method_union_and_inter}. Here, we conduct experiments to discuss the differences between these two methods. For the ``intersection'' method, we also use GPT-J to sample covariates with ``There are temporally ordered events [$E_i, E_{i + 1}, \ldots, E_n$]. Before all events,'' being the prompt template.

As shown in \cref{tab:part_union_inter}, the ``union'' method gets better performance since it can diversify covariate sets. It samples covariates conditioned on each event before $E_i$ separately. Meantime, each covariate of the ``intersection'' method is only conditioned on the same context $E_i, E_{i+1}, \ldots, E_n$. We compute the self-BLEU~\cite{zhu2018texygen} to evaluate the diversity of the generated covariates. The self-BLEU of the ``intersection'' method is 66.40\% while that of the ``union'' method is 41.34\%, quantitatively showing that our method can diversify the covariate set.
\begin{table}[t]
    \small
	\centering
	\begin{tabular}{l|ccc}
	\toprule
	\textbf{Models}&\textbf{Acc}&\textbf{F1}&\textbf{Ma-F1}\\
\midrule
ChatGPT&75.83&60.27&71.45\\
ChatGPT w/o $k$ &73.33&62.79&71.01\\
\modelname(DeBERTa-large) &\textbf{80.00}&\textbf{65.71}&\textbf{75.80}\\
\bottomrule
\end{tabular}
\caption{Comparison with ChatGPT. Our model outperforms ChatGPT with much fewer parameters.}
 \label{tab:chatgpt}
\end{table}

\subsection{Comparison with ChatGPT}
\label{sec:comparison_with_chatgpt}
Large language models have shown strong performance on extensive NLP tasks~\cite{radford2019language, ouyang2022training}. Thus, we compare our framework with ChatGPT\footnote{https://openai.com/blog/chatgpt/}, the latest large language model trained using Reinforcement Learning from Human Feedback (RLHF). To adopt ChatGPT to our task, we design the prompt template: ``Given a story as a chain of events $E_1, E_2, \ldots, E_n$, which $k$ event(s) in the first $n-1$ events more plausibly cause the last event?'' where $k$ is the number of positive events in the ground truth. We randomly sample 120 examples within 30 event sequences from our dataset \dataname and manually read predicted labels from ChatGPT's answers. We also test ChatGPT without providing the number $k$, i.e., removing $k$ from the prompt.

The experimental result in the selected 120 examples is shown in \cref{tab:chatgpt}. From the table, we can find that providing ChatGPT with the number $k$ does not lead to too much change in all metrics. (More discussion about this in \cref{sec:chatgpt_discussion}) Also, our framework \modelname achieves better zero-shot performance than ChatGPT with much fewer parameters.

\section{Conclusion}
In this paper, we design a new task to consider the context when detecting commonsense causal relations. We also crowd-sourced a dataset with strict quality control to benchmark the task. Our \modelname framework is motivated by the principles of causal inference and attempts to leverage context information. The extensive evaluation demonstrates
the effectiveness of \modelname for
detecting commonsense causal relations.

\section{Acknowledgement}
The authors of this paper were supported by the NSFC Fund (U20B2053) from the NSFC of China, the RIF (R6020-19 and R6021-20) and the GRF (16211520 and 16205322) from RGC of Hong Kong, the MHKJFS (MHP/001/19) from ITC of Hong Kong and the National Key R\&D Program of China (2019YFE0198200) with special thanks to HKMAAC and CUSBLT. 
This paper was also supported by the Tencent Rhino-bird Focused Research Program.
We also thank the support from NVIDIA AI Technology Center (NVAITC) and the UGC Research Matching Grants (RMGS20EG01-D, RMGS20CR11, RMGS20CR12, RMGS20EG19, RMGS20EG21, RMGS23CR05, RMGS23EG08).

\newpage
\section*{Limitations}
Our \modelname framework is mainly based on pre-trained language models, such as GPT-J and \texttt{PolyJuice}. We only utilize the simplest prompt to sample covariates and interventions. Further efforts such as prompt engineering are expected to boost the performance of \modelname further. Also, pre-trained language models can be biased by implicit events and reporting biases in their training data. Such biases lead the framework to omit some critical covariates and interventions, hindering our framework from achieving more accurate detection of commonsense causal relations. How to account for this problem of language model remains to be studied. Last but not least, the temporal propensity used in our framework is only an approximation of the exact propensity~\cite{zhang2022rock} since it is unclear how to pack
an unordered set of covariates into a sequential input for language models. Further studies are needed to design the exact propensity and improve performance.

\section*{Ethics Statement}
This work presents \dataname, a free and open dataset for the research community to study the contextualized CCR problem. Examples in \dataname are collected from ROCStories~\cite{mostafazadeh2016corpus}, a free and open dataset about anonymized chains of events. Each chain logically follows everyday topics and does not involve privacy problems about any specific entities (e.g., a person or company). We carried out human annotation on Amazon Mechanical Turk. Annotators are fairly paid 1.2 USD for each HIT, which fulfills the minimum wage requirement.

\newpage
\bibliography{anthology,custom}
\bibliographystyle{acl_natbib}

\appendix

\newpage
\section{Implementation Details}
In this appendix, we introduce the implementation details of every component in our framework \modelname. We conduct all experiments on 8 NVIDIA A100 GPUs. Since our \modelname and baselines are zero-shot, we test each model once and report the results.

\paragraph{Parameter Number} We list parameter numbers of all pre-trained langauge models here. BERT-base/large, RoBERTa-base/large, and DeBERTa-base/large have 110M, 340M, 125M, 355M, 100M, and 350M parameters, respectively. GPT, GPT-medium/large/XL, and GPT-6b contain 117M, 345M, 774M, 1.5B, and 6B parameters, respectively.

\subsection{Temporal Predictor}
\label{app:temporal_predictor}
We fine-tune temporal predictors based on BERT-base/large, RoBERTa-base/large, and DeBERTa-
base/large for ten epochs. The learning rate is 1e-5, and the batch size is 256. We sampled 800k examples (event pairs) from ROCStories~\cite{mostafazadeh2016corpus} to build a temporal relation dataset. The ROCStories dataset contains about 100k temporally ordered event sequences. The proportion of training, validation, and testing splits of our sampled event pairs are 98:1:1. We show the accuracy of each fine-tuned temporal predictor on the validation and testing splits in \cref{tab:temporal_accuracy}.
\begin{table}[t]
    \small
	\centering
	\begin{tabular}{l|cc}
	\toprule
	\textbf{Models}&\textbf{Valid Acc}&\textbf{Test Acc}\\
        \midrule
        Ours(BERT-base)&87.10&85.81 \\
        Ours(BERT-large)&88.93&87.38 \\
        Ours(RoBERTa-base)&89.76&88.97 \\
        Ours(RoBERTa-large)&91.91&91.10 \\
        Ours(DeBERTa-base)&90.07&89.06 \\
        Ours(DeBERTa-large)&93.10&92.40 \\
	\bottomrule
	\end{tabular}
	\caption{The accuracy of each temporal predictor on the validation and testing splits of the temporal relation dataset.}
        \label{tab:temporal_accuracy}
\end{table}

\subsection{Intervention and Covariate}
\label{app:controle_code}
For each event pair, we sample 50 covariates using GPT-J with a maximum length of generated events of 15 and a temperature of 0.9. We also sample 50 interventions using \texttt{PolyJuice} with a maximum length of generated events of 40 and a temperature of 1.0.

\texttt{PolyJuice} provided various control codes to manipulate events in various manners. The full list of control codes is \texttt{negation}, \texttt{lexical}, \texttt{resemantic},
    \texttt{quantifier}, \texttt{insert}, \texttt{restructure}, \texttt{shuffle},
    and \texttt{delete}. We only use \texttt{resemantic}, \texttt{negation}, \texttt{lexical}, \texttt{quantifier}, \texttt{insert}, \texttt{delete} to generate interventions since \texttt{restructure} and \texttt{shuffle} do not generate counterfactual events.

\subsection{Temporal Propensity Threshold $\epsilon$}
\label{sec:hyperparameter}
In this section, we list the best temporal propensity threshold $\epsilon$ for our framework \modelname with the temporal predictor based on different language models in \cref{tab:best_eps}.
\begin{table}[t]
    \small
	\centering
	\begin{tabular}{l|c}
	\toprule
	\textbf{Models}&\textbf{Best $\epsilon$}\\
        \midrule
        Ours(BERT-large)& 0.006 \\
        Ours(RoBERTa-large)& 0.001 \\
        Ours(DeBERTa-large)& 0.014 \\
	\bottomrule
	\end{tabular}
	\caption{The best $\epsilon$ for our framework \modelname with the temporal predictor based on different language models.}
        \label{tab:best_eps}
\end{table}

\begin{figure}[t]
    \centering
    \includegraphics[width=\columnwidth]{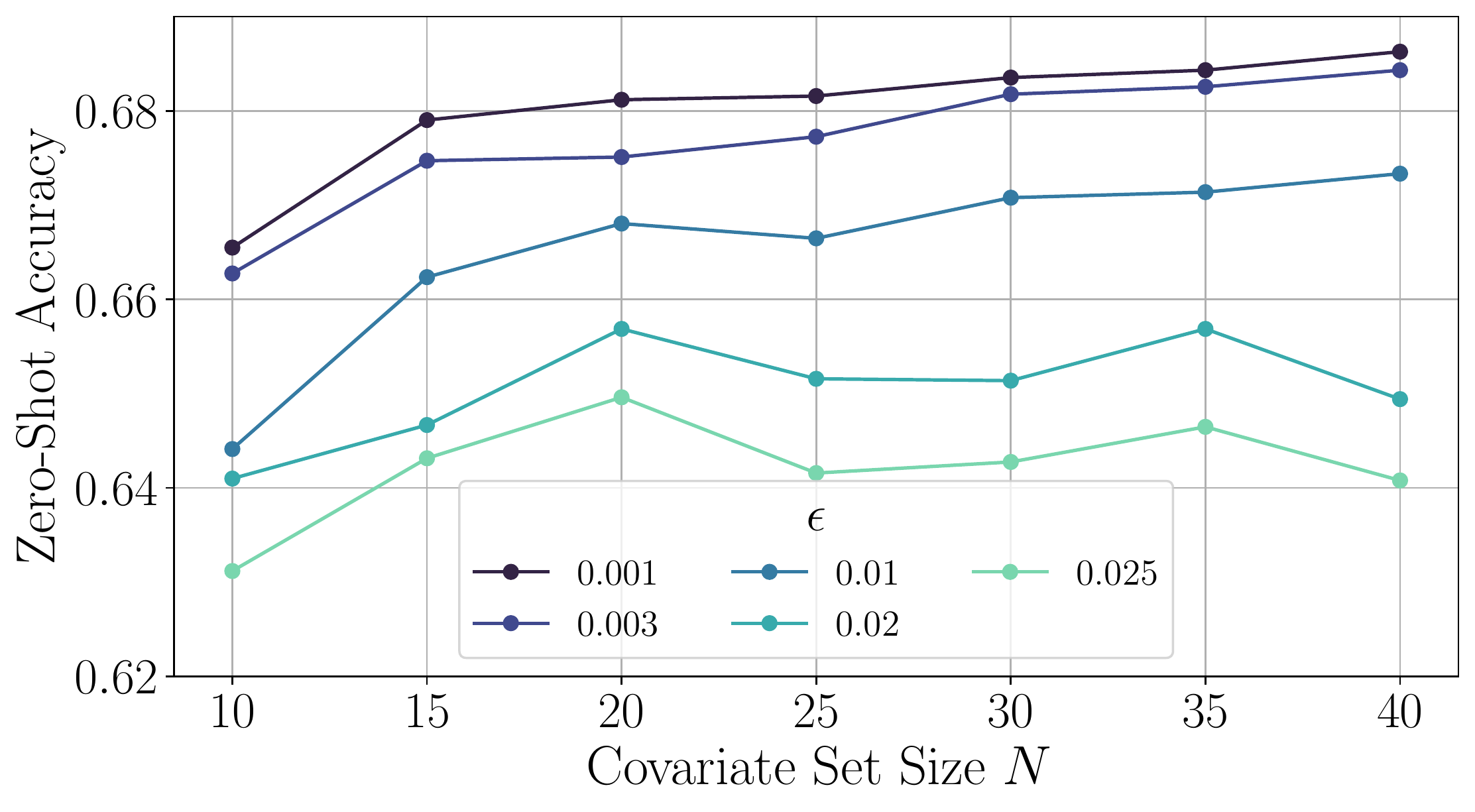}
    \caption{Accuracy of \modelname (``Ours'') with different covariate set size $N$. Increasing covariate set size improves performance with a reasonable $\epsilon$, such as 0.001, 0.003, and 0.01.}
    \label{fig:cov}
\end{figure}

\section{Probability Approximation}
\label{app:probability_approximation}
Our approximation of the probability of a covariate $\Pr(X(0))$ in \cref{sec:score_estimator} is a consequence of the efficiency trade-off.
Notice that even though a covariate $\sfX$ is an event preceding $\sfE_i$ sampled by GPT-J, it can occur before other events. Strictly, we must enumerate all possible events $\sfE$, compute $\Pr(\sfX \prec \sfE)$, and marginalize out $\sfE$. Here, we assume $\Pr(E(1)\vert X(0))\approx1$ to compute $\Pr(X(0))$ efficiently. We approximately have:
\begin{equation}
    \begin{aligned}
        \Pr(\sfX \prec \sfE) &= \Pr(X(0), E(1))\\
        &= \Pr(X(0)) \Pr(E(1)\vert X(0))\\
        &\approx \Pr(X(0))
    \end{aligned}
\end{equation}

\section{Number of Covariates}
\label{app:number_of_covariates}
We show that balancing covariates are essential in our framework \modelname. Meanwhile, the number of covariates also has an impact on performance. Here we test different covariate set size $N = \abs{\calX}$ spanning from 10 to 40, shown in \cref{fig:cov} with various $\epsilon$ values. 

We observe that gradually increasing the covariate set size $N$ can enhance the performance of our framework when the $\epsilon$ is within [0.001, 0.01] as we recommend in \cref{sec:choosing_epsilon}. On the other hand, adding more covariates may introduce more noise if we test with a larger $\epsilon$. For example, when $\epsilon$ equals 0.02 and 0.025, increasing covariates only brings about fluctuations in accuracy.

\section{Dataset Annotation}
\label{app:dataset_annotation}
Since our work is the first attempt to study Contextualized CCR, we carried out human annotation on Amazon Mechanical Turk to construct a new dataset. In this appendix, we discuss the annotation process of the dataset \dataname in our paper.
We randomly sampled event sequences from ROCStories~\cite{mostafazadeh2016corpus}, where each sequence contains five chronologically ordered events. Workers are provided with event sequences and are asked to annotate whether an event causes the last event in a sequence\footnote{There may be multiple causes of an event.}. Each human intelligence task (HIT) includes ten potential (\emph{cause}, \emph{effect}) event pairs (with corresponding event sequences), and each pair is labeled by seven workers. We take the majority vote among seven votes as the final result for each pair.

We conduct two qualification tests to choose workers to maintain rigorous quality control. First, we invited annotators who meet the following conditions to take our qualification examinations: 1) an approval rate of at least 95\% and 2) at least a thousand approved HITs. In the second round, a qualification question set including both effortless and tricky examples is collected by experts, namely the authors of this paper, who have a clear understanding of Contextualized CCR. The experts annotate 160 event pairs sampled from ROCStories. An annotator needs to answer a HIT involving ten questions from the qualification set, and the answers are compared with the experts' answers. An annotator should correctly answer 8 out of 10 questions to pass the second round test. While 307 annotators participated in the second round qualification test, only 29 (9.45\%) were selected as our main round annotators.

\begin{figure*}[t]
    \centering
    \includegraphics[width=1.6\columnwidth]{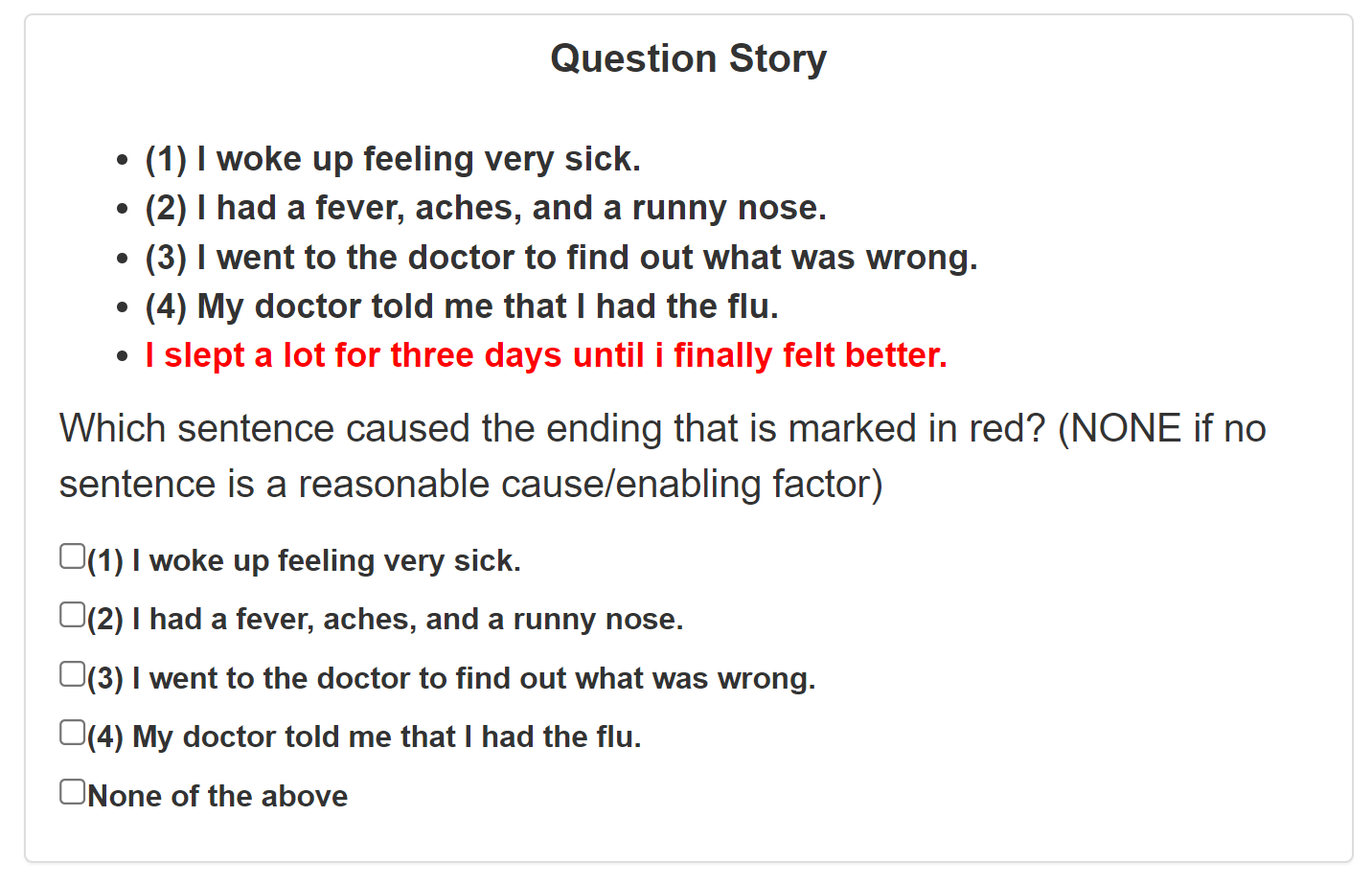}
    \caption{A screenshot of our annotation platform.}
    \label{annotation_screenshot}
\end{figure*}

Eventually, we collected the dataset containing human-annotated labels for 1,360 pairs from 340 event sequences ($1360 = 340\times4$), called \emph{Choice of Plausible Event in Sequence} (\dataname). The IAA score is 61.13\% calculated using pairwise agreement proportion, and Fleiss’s $\kappa$~\cite{fleiss1971measuring} is 0.52.  We equally divide them into a validation set and a testing set (e.g., each contains 680 examples). We also provide a screenshot of our annotation platform in \cref{annotation_screenshot}.                                                                                                                                                                                                                                                                                                                                                                                                                                  

\subsection{Dataset Statistics}
Here we provide the breakdown of numbers of causal relations $k$ in all 340 event sequences in \cref{tab:dataset_breakdown}. Also, 476 event pairs are positive (with causal relation), and 884 are negative (without causal relation).

\begin{table}[t]
    \small
	\centering
	\begin{tabular}{l|ccccc}
	\toprule
	\textbf{\#Causal Relation}&\textbf{0}&\textbf{1}&\textbf{2}&\textbf{3}&\textbf{Total}\\
\midrule
\textbf{\#Seq }&12&192&124&12&340\\
\midrule
\textbf{Proportion (\%)}&3.5&56.5&36.5& 3.5&100\\
\bottomrule
\end{tabular}
\caption{The breakdown of numbers of causal relations in all event sequences in the \dataname dataset. Most event sequences (56.6\%) own only one causal relation.}
 \label{tab:dataset_breakdown}
\end{table}

\section{Normalization}
\label{sec:normalization}
In this section, we enumerate some methods to normalize estimand $\hat{\Delta}$ and temporal propensity vectors in \cref{eq:estimated_ate}.
\\

\noindent\textbf{Direct Matching ($\mathbf{D}$):} Instead of condition probability in \cref{eq:est_propensity}, we can directly use the vectors of temporal relation scores $\left(f(\sfA, \sfX)\right)_{\sfX\in\calX}$ as propensity vectors.
\\

\noindent\textbf{Score Simplification ($\mathbf{S}$):} We fine-tune temporal predictors to prompt \texttt{before}, \texttt{after}, and \texttt{[none]}, which might be difficult for smaller Masked LMs, like BERT-base. When constructing propensity vectors, we simplify the task to consider only \texttt{before}, \texttt{after} by normalizing scores: 
\begin{equation*}
    \begin{aligned}
    f(\sfX, \sfA)= \resizebox{0.78\columnwidth}{!}{$\frac{f_b(\sfX, \sfA) + f_a(\sfA, \sfX)}{f_b(\sfX, \sfA) + f_a(\sfX, \sfA) + f_b(\sfA, \sfX) + f_a(\sfA, \sfX)}$}
    \end{aligned}
\end{equation*}

\noindent\textbf{Propensity Covariate Normalization ($\mathbf{Q}$):} We also try to normalize temporal relation scores on the covariate set $\calX$ before
building propensity vectors: $\Pr(X) = f(\sfX, \sfE_i) / \sum_{\sfX'\in\calX} f(\sfX', \sfE_i)$
    and $\Pr(X, A) = f(\sfX, \sfA)/ \sum_{\sfX'\in\calX} f(\sfX', \sfA)$.
\\
 
\noindent\textbf{Co-occurrence Normalization ($\mathbf{C}$):} The fine-tuned temporal predictor may sometimes still faces the problem of low coverage, causing estimates $f(\sfE_i, \sfE_n)$ and $f(\sfA, \sfE_n)$ inaccurate in \cref{eq:estimated_ate}. We set them to $(f(\sfE_i, \sfE_n) + f(\sfE_n, \sfE_i)) / 2$ and $(f(\sfA, \sfE_n) + f(\sfE_n, \sfA)) / 2$, respectively.
\\

\noindent\textbf{Estimand Normalization ($\mathbf{E}$):}
In this method, estimates $f(\sfE_i, \sfE_n)$ and $f(\sfA, \sfE_n)$ in estimand $\hat{\Delta}$ in \cref{eq:estimated_ate} are normalized by $f(\sfE_i, \sfE_n) + f(\sfE_n, \sfE_i)$ and $f(\sfA, \sfE_n) + f(\sfE_n, \sfA)$, respectively.

We also conduct comprehensive analyses about removing these normalizations. The results are shown in \cref{tab:normal_bert,tab:normal_roberta,tab:normal_deberta}. We present drops in every metric when removing each normalization method. Some normalization methods cannot improve performance, and we delete those rows from tables. We observe that some of these normalization methods, such as \textbf{Estimand Normalization ($\mathbf{E}$)}, can benefit our framework on Contextualized CCR.

\begin{table}[t]
    \small
	\centering
	\begin{tabular}{l|ccc}
	\toprule
	\textbf{Models}&\textbf{Acc}&\textbf{F1}&\textbf{Ma-F1}\\
        \midrule
        Ours(BERT-base)&68.82&55.27&65.67 \\
        \midrule
        $\diamond$ w/o S&$\downarrow$ 0.59&$\downarrow$ 0.84&$\downarrow$ 0.65 \\
        $\diamond$ w/o Q&$\downarrow$ 3.24&$\downarrow$ 4.64&$\downarrow$ 3.56 \\
        $\diamond$ w/o E&$\downarrow$ 5.29&$\downarrow$ 7.59&$\downarrow$ 5.83 \\
        \midrule \midrule
        Ours(BERT-large)&70.29&57.38&67.29 \\
        \midrule
        $\diamond$ w/o Q&$\downarrow$ 0.59&$\downarrow$ 0.84&$\downarrow$ 0.65 \\
        $\diamond$ w/o E&$\downarrow$ 7.06&$\downarrow$ 10.13&$\downarrow$ 7.77 \\
        \bottomrule
	\end{tabular}
	\caption{Ablation study of normalization methods on BERT-base and BERT-large.}
        \label{tab:normal_bert}
\end{table}

\begin{table}[t]
    \small
	\centering
	\begin{tabular}{l|ccc}
	\toprule
	\textbf{Models}&\textbf{Acc}&\textbf{F1}&\textbf{Ma-F1}\\
        \midrule
        Ours(RoBERTa-base)&66.76&52.32&63.41 \\
        \midrule
        $\diamond$ w/o C&$\downarrow$ 1.47&$\downarrow$ 2.11&$\downarrow$ 1.62 \\
        \midrule \midrule
        Ours(RoBERTa-large)&70.00&56.96&66.97 \\
        \midrule
        $\diamond$ w/o S&$\downarrow$ 0.59&$\downarrow$ 0.84&$\downarrow$ 0.65 \\
        $\diamond$ w/o E&$\downarrow$ 2.06&$\downarrow$ 2.95&$\downarrow$ 2.27 \\
        \bottomrule
	\end{tabular}
	\caption{Ablation study of normalization methods on RoBERTa-base and RoBERTa-large.}
        \label{tab:normal_roberta}
\end{table}
\begin{table}[t]
    \small
	\centering
	\begin{tabular}{l|ccc}
	\toprule
	\textbf{Models}&\textbf{Acc}&\textbf{F1}&\textbf{Ma-F1}\\
        \midrule
        Ours(DeBERTa-base)&70.29&57.38&67.29 \\
        \midrule
        $\diamond$ w/o Q&$\downarrow$ 0.59&$\downarrow$ 0.84&$\downarrow$ 0.65 \\
        \midrule \midrule
        Ours(DeBERTa-large)&69.41&56.12&66.32 \\
        \midrule
        $\diamond$ w/o E&$\downarrow$ 1.18&$\downarrow$ 1.69&$\downarrow$ 1.30 \\
	\bottomrule
	\end{tabular}
	\caption{Ablation study of normalization methods on DeBERTa-base and DeBERTa-large.}
        \label{tab:normal_deberta}
\end{table}

\section{ChatGPT}
\label{sec:chatgpt_discussion}
\newcommand{\shortline}{------------------------------\\}

In this section, we show and discuss the result of ChatGPT on \dataname. Given the insignificant differences in performance between the prompts with and without the number $k$ in \cref{sec:comparison_with_chatgpt}, we focus on the answers of the prompt with $k$. We show three examples, respectively corresponding to three cases:
\begin{enumerate}
\item[1.] ChatGPT's answer is correct (with a good explanation)
\item[2.] ChatGPT's answer is different from the ground truth answer, but the correctness is debatable with the explanation from ChatGPT
\item[3.] ChatGPT's answer is incorrect.
\end{enumerate}

\noindent
\textbf{Example 1.} Which event mainly/directly causes the last event? \\
{\small
$\sf{E}_1$: Charlie went to food truck event at the beach. \\
$\sf{E}_2$: He said they had a ton of trucks there. \\
$\sf{E}_3$: He tried food from a few trucks. \\
$\sf{E}_4$: He said the food was great. \\
$\sf{E}_5$: I am going to try to go next year. \\
\shortline
\textbf{Gold answer}: $\sf{E}_4$ \\
\textbf{ChatGPT answer (short)}: $\sf{E}_4$ \\
\textbf{ChatGPT answer (original)}: The event "He said the food was great." directly causes the last event "I am going to try to go next year." as it expresses the protagonist's positive experience and desire to attend again in the future. \\
}

\noindent
\textbf{Example 2.} Which event mainly/directly causes the last event? \\
{\small
$\sf{E}_1$: Ericka enjoyed going to restaurants, but she liked to save money too. \\
$\sf{E}_2$: She found a coupon for her favorite restaurant. \\
$\sf{E}_3$: She went there for lunch and ordered a burger. \\
$\sf{E}_4$: When the bill came, she was nervous about the coupon. \\
$\sf{E}_5$: The restaurant accepted the coupon and Ericka saved a lot of money! \\
\shortline
\textbf{Gold answer}: $\sf{E}_2, \sf{E}_3$ \\
\textbf{ChatGPT answer (short)}: $\sf{E}_2, \sf{E}_4$ \\
\textbf{ChatGPT answer (original)}: The second and fourth events mainly/directly cause the last event. The second event ("She found a coupon for her favorite restaurant.") sets up the possibility of her saving money, and the fourth event ("When the bill came, she was nervous about the coupon.") establishes that the coupon is being used in this specific instance. Together, these two events lead to the final event ("The restaurant accepted the coupon and ericka saved a lot of money!") where she is able to save money. \\
}

\noindent
\textbf{Example 3.} Which event mainly/directly causes the last event? \\
{\small
$\sf{E}_1$: We decided to get married on the beach. \\
$\sf{E}_2$: Our family and friends met us there. \\
$\sf{E}_3$: It was a perfectly sunny summer day. \\
$\sf{E}_4$: Best of all, no one had to wear shoes!. \\
$\sf{E}_5$: We ended our day with a dip in the cool ocean. \\
\shortline
\textbf{Gold answer}: $\sf{E}_1$ \\
\textbf{ChatGPT answer (short)}: $\sf{E}_4$ \\
\textbf{ChatGPT answer (original)}: The fourth event "Best of all, no one had to wear shoes!" mainly/directly causes the last event "We ended our day with a dip in the cool ocean." as it describes the condition that allowed the last event to happen: people were barefoot. \\
}

\noindent
In most cases, ChatGPT clearly explains its choices. For example, in the \textbf{Example 1}, it points out the positive \textbf{emotion} \textit{``... food was great.''} as the underlying fact that causes the last event \textit{``... going to try to go next year.''} (i.e., \textbf{decision}). Similarly in the \textbf{Example 2}, even though ChatGPT's answer is different from the ground truth answer, ChatGPT mentions two key points \textit{``... The second event ... sets up the possibility of ...''} (i.e., \textbf{potential}) and \textit{``.. the forth event ... establishes that the coupon is being used ...''} (i.e., \textbf{action}) that together causes the last event (i.e., \textbf{effect}). The reasoning paradigm implicit in its explanation, e.g \textbf{emotion $\to$ decision} and \textbf{potential + action $\to$ effect} in the abovementioned two examples, implies the inherent reasoning capability of ChatGPT. Nonetheless, there are cases where ChatGPT fails to give a persuasive answer. However, as a foundation model instead of a task model, ChatGPT's performance is fair enough.

\end{document}